\documentclass[runningheads]{llncs}
\usepackage[T1]{fontenc}
\usepackage{amsmath,amssymb,amsfonts}
\usepackage{mathtools}
\usepackage{array}
\usepackage{booktabs}
\usepackage{graphicx}
\usepackage{microtype}
\usepackage{xcolor}
\usepackage{rotating}
\usepackage{multirow}
\usepackage{url}
\usepackage[hidelinks]{hyperref}

\newcommand{\Ebase}{E_{\mathrm{b}}}
\newcommand{\Ekeep}{\widetilde{E}}
\newcommand{\Eopt}{E^\star}
\newcommand{\Gbase}{G_{\mathrm{b}}}
\newcommand{\Gkeep}{\widetilde{G}}
\newcolumntype{C}[1]{>{\centering\arraybackslash}m{#1}}

\renewcommand{\arraystretch}{1.02}
\begin{document}

\title{Machine Learning-based Two-Stage Graph Sparsification for the Travelling Salesman Problem}
\titlerunning{ML-based Two-Stage Graph Sparsification for TSP}
\author{Bo-Cheng Lin, Yi Mei, Mengjie Zhang}
\authorrunning{Bo-Cheng Lin, Yi Mei, Mengjie Zhang}
\institute{
Centre for Data Science and Artificial Intelligence, 
\\ and the School of Engineering and Computer Science, 
\\Victoria University of Wellington, Wellington, 6140, New Zealand 
\\ \{ bocheng.lin, yi.mei, mengjie.zhang \} @ecs.vuw.ac.nz}

\maketitle

\begin{abstract}

High-performance TSP solvers such as Lin-Kernighan-Helsgaun (LKH) search within a \emph{candidate graph}---a small subset of edges pre-selected for the solver---rather than over the complete graph. The two leading sparsification heuristics, $\alpha$-Nearest and POPMUSIC, each fall short of the density--coverage balance: $\alpha$-Nearest is dense with stable recall, while POPMUSIC is sparser but its recall degrades with scale. Their union closes the recall gap while remaining far below the complete graph in density, leaving room for further reduction. Existing learning-based sparsifiers score edges on the complete graph, an approach that is expensive and largely limited to Euclidean instances. We propose a two-stage method that inverts this logic. Stage~1 takes the union of $\alpha$-Nearest and POPMUSIC, achieving near-perfect recall at ${\sim}6N$ edges. Crucially, the union annotates each edge with its \emph{source provenance}---whether it was endorsed by $\alpha$-Nearest, POPMUSIC, or both. Stage~2 trains a lightweight classifier on these annotated edges and prunes the lowest-scoring ones. Because dual-source edges are almost always optimal, the learning problem reduces to filtering the single-source subset---a substantially easier task than classifying all $O(N^2)$ edges from scratch. Across four distance types, five spatial distributions, and problem sizes from 50 to 500, the pipeline reduces candidate-graph density by $37$--$47\%$ while retaining ${\geq}99.69\%$ of optimal-tour edges, and matches or exceeds the coverage of recent Euclidean-only neural sparsifiers at lower density at TSP500.
\end{abstract}

\section{Introduction}
\vspace{-1mm}
High-performance travelling salesman problem (TSP) solvers do not try all possible edges. Instead, they work within a \emph{candidate graph}, a small subset of edges that the solver is allowed to use. Heins et al.\ \cite{heinsDancingStateArt2024a} shows that candidate-graph design is a first-order determinant of LKH performance, and candidate graphs are equally important in other frameworks such as ant colony optimisation \cite{randallCandidateSetStrategies2002}. A good candidate graph must be sparse enough for fast search but must keep the edges that belong to high-quality tours. No existing method fully achieved this trade-off.

The two sparsification heuristics integrated into LKH~\cite{helsgaun2000effective}, $\alpha$-Nearest and POPMUSIC, are also the strongest in the recent benchmark of Heins et al.~\cite{heinsDancingStateArt2024a}. Individually, each heuristic leaves a gap: $\alpha$-Nearest is dense (${\sim}6N$ edges) with stable recall; POPMUSIC is sparser (${\sim}1$--$2N$) but its recall degrades at larger scales. Although they cannot be effective individually, their union can keep almost all the promising edges with a relatively high density (${\sim}6N$), but still much sparser than a complete graph ($N(N{-}1)/2$). This leaves an open question: \emph{can we further reduce this density while preserving the near-perfect recall that the union achieved?}

We propose a \textbf{two-stage} approach: first reduce as many edges as possible while maximising recall, then further reduce density. Stage~1 takes the union of $\alpha$-Nearest and POPMUSIC to build a candidate graph with near-perfect coverage. Stage~2 trains a machine learning model to score each edge and prune the lowest-scoring ones. Two properties make this pipeline effective. First, $\alpha$-Nearest and POPMUSIC have complementary failure modes: $\alpha$-Nearest ranks edges by 1-tree structure and misses long chords, while POPMUSIC builds edges from local-tour optima and misses moderate-length edges crossing subproblem boundaries; their union absorbs both kinds of misses, so Stage~1 alone already achieves ${\geq}99.96\%$ recall. Second, the union annotates every edge with its source provenance---whether it was endorsed by $\alpha$-Nearest, POPMUSIC, or both. Dual-source edges are almost always optimal, so the learning problem in Stage~2 reduces to identifying which single-source edges to discard---a substantially easier task than classifying all $N(N{-}1)/2$ edges from scratch. Reversing the order would forfeit both benefits. All features depend only on edge weights, so the method is \emph{distance-type agnostic}. Following multi-metric evaluation \cite{kerschkeParameterizationStateoftheartPerformance2018}, we report density, coverage, and downstream solver performance jointly. Our contributions are:

\begin{itemize}
\item We propose a two-stage graph sparsification pipeline that applies learned pruning to a sparse candidate graph rather than to the complete graph. A single model, trained once on a mixed dataset of 20 TSP families (4 distance types $\times$ 5 spatial distributions), reduces density by $37$--$47\%$ while retaining ${\geq}99.69\%$ of optimal-tour edges. On Euclidean instances, the method matches or exceeds the coverage of recent neural sparsifiers at lower density.
\item We show that fusing two heuristics ($\alpha$-Nearest $\cup$ POPMUSIC) creates a source-provenance signal that makes learned pruning effective: even a simple logistic regression suffices, and the method transfers stably to instances five times larger than the training size.
\item We demonstrate that the two-stage method becomes increasingly valuable at larger scales: from $N{=}200$ onward, union + learned pruning achieves higher coverage than POPMUSIC alone, even though POPMUSIC is the stronger single-stage heuristic at small sizes.
\end{itemize}
\vspace{-1mm}

\vspace{-3mm}
\section{Background and Related Work}
\vspace{-2mm}
\subsection{Problem Definition}
\vspace{-1mm}

The Travelling Salesman Problem (TSP) is an $\mathcal{NP}$-hard combinatorial optimisation problem defined on a complete undirected graph $G=(V,E)$ with vertex set $V=\{1,\ldots,N\}$, edge set $E=\{\{i,j\}:i,j\in V,\,i\neq j\}$, and a edge-cost function $c:E\to\mathbb{R}_{\geq 0}$, where $d_{ij}\coloneqq c(\{i,j\})$ denotes the cost of edge $\{i,j\}$. A \emph{tour} is a Hamiltonian cycle on $G$, that is, a closed walk visiting every vertex exactly once, whose edge set $\tau\subseteq E$ satisfies $|\tau|=N$. The TSP asks for an optimal tour $\tau^{\star}\in\arg\min_{\tau}\sum_{e\in\tau}c(e)$, and we write $\Eopt=\tau^{\star}$ for its edge set. 

\vspace{-3mm}
\subsection{Sparsification Heuristics}\label{sec:background-candidates}
\vspace{-1mm}
A \emph{candidate graph} $G_c=(V,E_c)$ with $E_c\subseteq E$ restricts which edges the solver may explore; edges outside $E_c$ are never considered. The quality of a candidate graph is measured by its \emph{coverage} (the fraction of $\Eopt$ retained, $|E_c\cap\Eopt|/N$) and its \emph{density} (the number of edges per node, $|E_c|/N$). The complete graph trivially achieves $100\%$ coverage at $N(N{-}1)/2$ edges; the challenge is to maintain near-perfect coverage while minimising density.

The two leading sparsification heuristics, both integrated in LKH \cite{helsgaun2000effective}, are:

\textbf{$\alpha$-Nearest.} Ranks edges by their $\alpha$-values derived from minimum 1-trees. Smaller $\alpha_e$ means the edge is structurally compatible with good tours. LKH keeps the five lowest-$\alpha$ edges per node, producing graphs of ${\sim}5$--$6N$ edges.

\textbf{POPMUSIC} \cite{taillardPopmusicPartialOptimization,taillardPOPMUSICTravellingSalesman2019}. Runs local tour improvement on overlapping subproblems and keeps every edge that appears in an improved local tour. This yields sparser graphs (${\sim}1$--$2N$ edges) but with less recall at large scales.

Both methods operate purely on edge weights, so they work for any distance convention. This sets them apart from Delaunay-based approaches \cite{dillencourtTravelingSalesmanCycles1987,krasnogorNewHybridHeuristic1999} and other triangulation methods \cite{letchfordGoodTriangulationsYield2008,tuononenModifiedGreedyDelaunay2024}, which require Euclidean coordinates.

Combinatorial rules can provably preserve optimal-tour edges~\cite{hougardyEdgeEliminationTSP2014}. Zhong~\cite{zhongProbabilisticAnalysisEdge2024} analysed such rules probabilistically, and Cook et al.~\cite{cookLocalEliminationTraveling2024} showed that local elimination can yield very sparse graphs on large instances. Fekete et al.~\cite{feketeEdgeSparsificationGeometric2024} and Wang~\cite{wangApproximateMethodCompute2015} studied geometric and frequency-based sparsification. These methods offer theoretical guarantees but are restricted to Euclidean distance and need long computation times.

\vspace{-3mm}
\subsection{Learning-Based TSP Methods}
\vspace{-1mm}
Machine learning has been widely applied to TSP. End-to-end neural solvers \cite{belloNeuralCombinatorialOptimization2017,koolAttentionLearnSolve2019b,joshiLearningTSPRequires2021,alanziSolvingTravelingSalesman2025} learn to construct tours directly, while hybrid methods integrate learning with classical solvers: NeuroLKH predicts edge guidance for LKH \cite{xinNeuroLKHCombiningDeep2021}, Hudson et al.\ combine graph neural networks with guided local search \cite{hudsonGraphNeuralNetwork2022}, Zheng et al.\ reinforce decisions inside LKH-style search \cite{zhengReinforcedLinKernighan2023}, and Wang et al.\ dynamically adapt candidate edges during search \cite{wangBanditBasedDynamic2025}. Learning has also been applied to exact-solver components such as branching and subtour-cut generation \cite{gasseExactCombinatorialOptimization2019,voImprovingSubtourElimination2023}.

The closest prior work applies learning to graph sparsification. GNN-based methods predict edge heatmaps \cite{sunDIFUSCOGraphbasedDiffusion2024,joshiEfficientGraphConvolutional2019}, and edge confidence scores can be viewed as implicit sparsifiers \cite{lischkaTravellingSalesmanProblem2023,lischkaLessMoreImportance2024}. Sun et al.\ \cite{sunUsingStatisticalMeasures2021,sunGeneralizationMachineLearning2021} introduced a machine-learning problem-reduction framework using statistical edge features and studied how such reductions generalise across TSP families. Fitzpatrick et al.\ \cite{fitzpatrickLearningSparsifyTravelling2021} learn to sparsify for exact formulations, and Zhou et al.\ \cite{zhouLearningReduceSearch2025} dynamically reduce the search space during autoregressive solving. However, all these methods operate on the \emph{complete graph} or dense subgraphs such as KNN-50 or KNN-100, which is expensive and scales poorly. They are also almost exclusively Euclidean. No prior work has applied learning to an already highly sparse candidate graph. Our own feature design builds on these statistical edge features and the broader TSP feature-engineering tradition \cite{heinsPotentialNormalizedTSP2021,heinsStudyEffectsNormalized2023} developed for algorithm selection \cite{kotthoffImprovingStateArt2015,kerschkeLeveragingTSPSolver2018}.

\vspace{-3mm}
\section{Method}
\vspace{-2mm}
\subsection{Pipeline Overview}
\vspace{-1mm}

\begin{figure}[!h]
\centering
\includegraphics[width=\columnwidth]{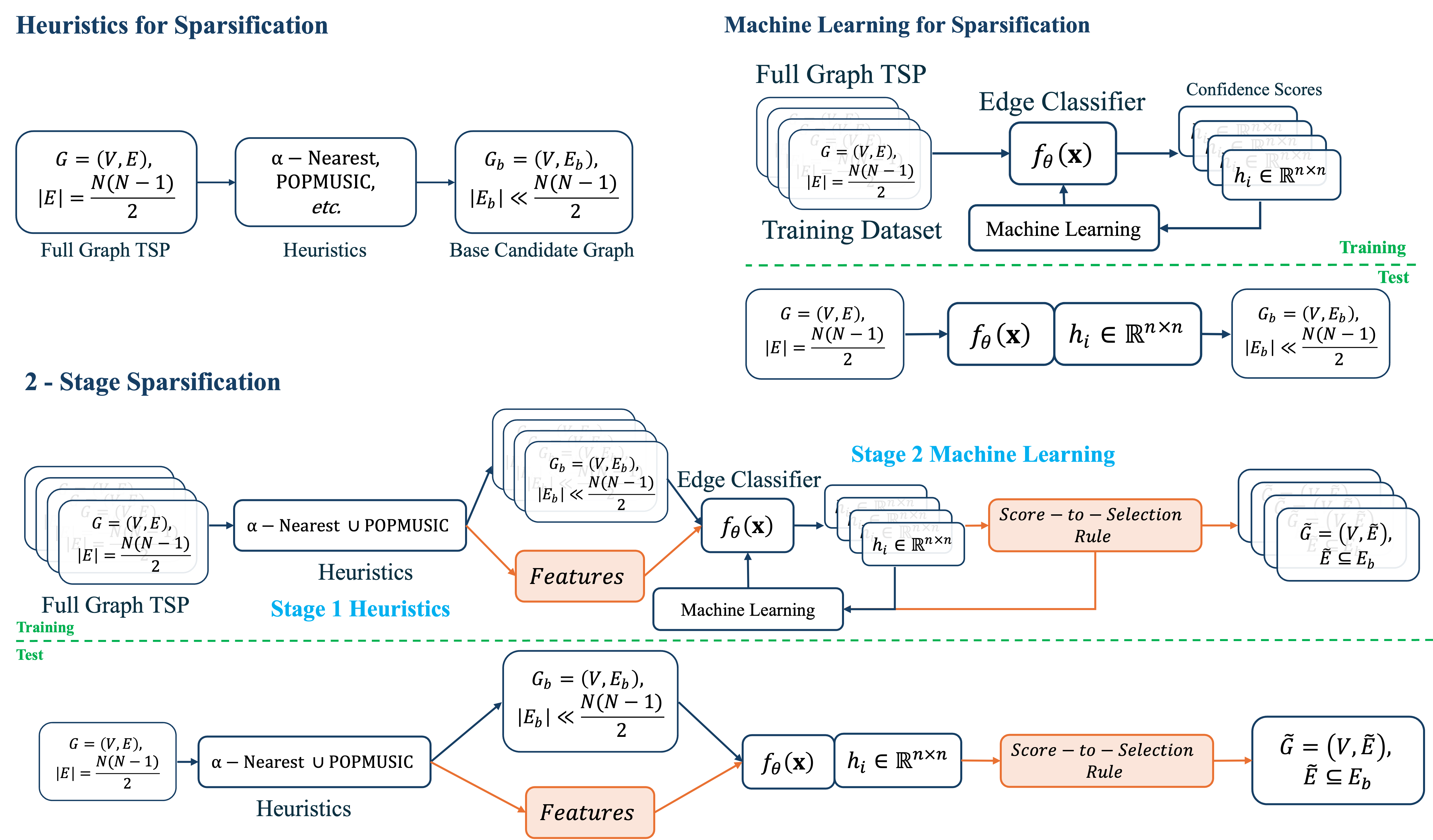}
\caption{Single-stage vs.\ two-stage sparsification. Top: existing approaches apply heuristics or ML independently to the full graph. Bottom: our method first reduces edges while preserving recall via a heuristic union, then reduces density further via learned pruning.}
\label{fig:flowchart}
\end{figure}

Figure~\ref{fig:flowchart} illustrates our two-stage approach. In \textbf{Stage~1}, a heuristic constructs a \emph{base candidate graph} $\Gbase = (V, \Ebase)$ with $|\Ebase| \ll N(N{-}1)/2$. We use the union of $\alpha$-Nearest and POPMUSIC (Section~\ref{sec:background-candidates}):
\vspace{-1mm}
\begin{equation}
E_{\mathrm{union}} = E_{\alpha} \cup E_{\mathrm{pop}}.
\end{equation}

\vspace{-2mm}

The union does two things at once: it reduces the least promising edges while maximising recall, and it annotates every candidate edge with its heuristic provenance, a signal that no single heuristic can provide. These two binary indicators (Table~\ref{tab:features_main}) flow from Stage~1 into Stage~2, which is the key information transfer that makes learned pruning effective.

In \textbf{Stage~2}, a learned model scores every edge in $\Ebase$ and removes the lowest-scoring ones, producing a \emph{pruned candidate graph} $\Gkeep = (V, \Ekeep)$ with $\Ekeep \subseteq \Ebase$. The design of Stage~1 means Stage~2 learns from ${\sim}6N$ edges with a substantially more balanced class distribution and with the source-provenance signal available---a smaller and easier learning problem than classifying all $N(N{-}1)/2$ edges from scratch.

\textbf{Training.} For each training instance, we run Stage~1 to obtain $\Ebase$, compute the optimal tour with Concorde \cite{applegate2006concorde}, and label each candidate edge as $y_e = \mathbf{1}[e \in \Eopt]$. We extract a feature vector $x_e \in \mathbb{R}^p$ for each edge and train a scoring model $f_\theta$ on these $(x_e, y_e)$ pairs. At test time, the model scores all edges in a new base candidate graph, and a local pruning rule converts the scores into $\Gkeep$.

\vspace{-3mm}
\subsection{Stage 2: Learned Pruning}\label{sec:stage2}
\vspace{-1mm}

\subsubsection{Edge Features.}\label{sec:features}

For each edge $e = \{i,j\} \in \Ebase$, we compute a feature vector from six families (Table~\ref{tab:features_main}). Our features extend those of Sun et al.~\cite{sunUsingStatisticalMeasures2021,sunGeneralizationMachineLearning2021}. Every feature is \emph{distance-type agnostic}: we use ranks, ratios, and local statistics rather than coordinates, so the features work across all distance types.

\begin{table*}[!h]
\caption{Edge features for the learned pruning pipeline, extended from Sun et al.~\cite{sunUsingStatisticalMeasures2021,sunGeneralizationMachineLearning2021} with source-provenance indicators.\label{tab:features_main}}
\vspace{-3mm}
\centering
\scriptsize
\setlength{\tabcolsep}{4pt}
\begin{tabular}{C{0.17\textwidth} C{0.35\textwidth} C{0.38\textwidth}}
\toprule
Feature family & Description & Formula or definition \\
\midrule
Distance magnitude & Raw edge distance & $d_{ij}=d(i,j)$ \\
\midrule
\multirow{4}{*}{Distance ranks} & Endpoint rank percentile from $i$ & $r_i(j)/(N{-}1)$ \\
 & Endpoint rank percentile from $j$ & $r_j(i)/(N{-}1)$ \\
 & Min endpoint rank percentile & $\min\{r_i(j),r_j(i)\}/(N{-}1)$ \\
 & Max endpoint rank percentile & $\max\{r_i(j),r_j(i)\}/(N{-}1)$ \\
\midrule
\multirow{4}{*}{\shortstack{Local\\normalisation}} & Ratio to nearest neighbour at $i$ & $d_{ij}/\min_{u\neq i} d(i,u)$ \\
 & Ratio to nearest neighbour at $j$ & $d_{ij}/\min_{u\neq j} d(j,u)$ \\
 & Distance $z$-score at $i$ & $(d_{ij}-\mu_i)/\sigma_i$ \\
 & Distance $z$-score at $j$ & $(d_{ij}-\mu_j)/\sigma_j$ \\
\midrule
\multirow{2}{*}{\shortstack{Neighbourhood\\structure}} & Mutual-$k$NN indicator & $\mathbf{1}[j\in\mathrm{kNN}(i)\wedge i\in\mathrm{kNN}(j)]$ \\
 & $k$NN overlap ratio & $|\mathrm{kNN}(i)\cap\mathrm{kNN}(j)|/|\mathrm{kNN}(i)\cup\mathrm{kNN}(j)|$ \\
\midrule
\multirow{3}{*}{\shortstack{Candidate\\topology}} & Degree of $i$ in $\Ebase$ & $\deg_{\Ebase}(i)$ \\
 & Degree of $j$ in $\Ebase$ & $\deg_{\Ebase}(j)$ \\
 & Common candidate neighbours & $|N_{\Ebase}(i)\cap N_{\Ebase}(j)|$ \\
\midrule
\multirow{2}{*}{\shortstack{Source\\provenance}} & Edge in $\alpha$-Nearest & $\mathbf{1}[e\in E_{\alpha}]$ \\
 & Edge in POPMUSIC & $\mathbf{1}[e\in E_{\mathrm{pop}}]$ \\
\bottomrule
\end{tabular}
\par\vspace{0.3em}
\noindent{\scriptsize All features depend only on edge weights and candidate-graph structure, not on coordinate geometry. Source-provenance features are available only in union mode. $r_i(j)$ is the rank of $d_{ij}$ among all distances from $i$; $\mu_i, \sigma_i$ are the mean and standard deviation of distances from $i$.}
\vspace{-1mm}
\end{table*}

\vspace{-2mm}
\subsubsection{Training Objective.}

Each model produces a scalar edge score $s_e = f_\theta(x_e)$ per edge. Logistic regression (LR) minimises weighted binary cross-entropy with $\ell_2$ regularisation; linear SVM minimises weighted squared-hinge loss; XGBoost minimises weighted binary cross-entropy over an additive tree ensemble. All three are lightweight, edge-independent classifiers. The choice of linear and tree-based models is deliberate: it lets the contributions of the feature design and the base-graph choice be read directly from the results (Section~\ref{sec:features_failures}), without the confounding capacity of a deep network. We deliberately avoid GNNs in this study because they demand high-performance GPUs, need substantially more training data, and would conflate representation learning with the pipeline design. Hyperparameters are listed in Section~\ref{sec:experiment_params}.

\vspace{-3mm}
\subsubsection{Score-to-Selection Rule.}\label{sec:pruning}

A global score threshold works poorly because node degrees in $\Gbase$ vary widely. We instead use a \emph{node-level} rule. For each node $i$, let $\delta(i)$ be its incident candidate edges sorted by descending score. We define a local softmax distribution:

\vspace{-1mm}

\begin{equation}
w_{i,e} 
= \mathrm{softmax}_{\,e \in \delta(i)}\!\left(\frac{s_e - s_i^{\max}}{T}\right)
\end{equation}

\vspace{-1mm}

where $s_i^{\max} = \max_{e \in \delta(i)} s_e$ and $T > 0$ is a temperature. We keep the top edges by descending score until their cumulative softmax mass reaches a threshold $\eta \in (0,1]$, with a hard floor of $m_{\min}{=}2$ edges per node to preserve tour feasibility. The threshold $\eta$ controls the sparsity-coverage trade-off: smaller $\eta$ retains fewer edges (sparser graph, lower coverage); larger $\eta$ retains more. Nodes with spread-out scores keep more edges; nodes with a clear winner keep fewer. The final graph $\Gkeep$ is the union of edges retained from both endpoints. This node-level softmax rule is, to our knowledge, a novel design for learned graph sparsification; the conventional alternative, a global score threshold, fails because node degrees in $\Gbase$ vary widely across the graph. Parameter values and selection procedure are in Section~\ref{sec:experiment_params}.

\vspace{-3mm}
\section{Experimental Design}

\vspace{-2mm}
\subsection{Benchmark}
\vspace{-1mm}

\begin{figure}[!h]
\centering
\includegraphics[width=0.9\columnwidth]{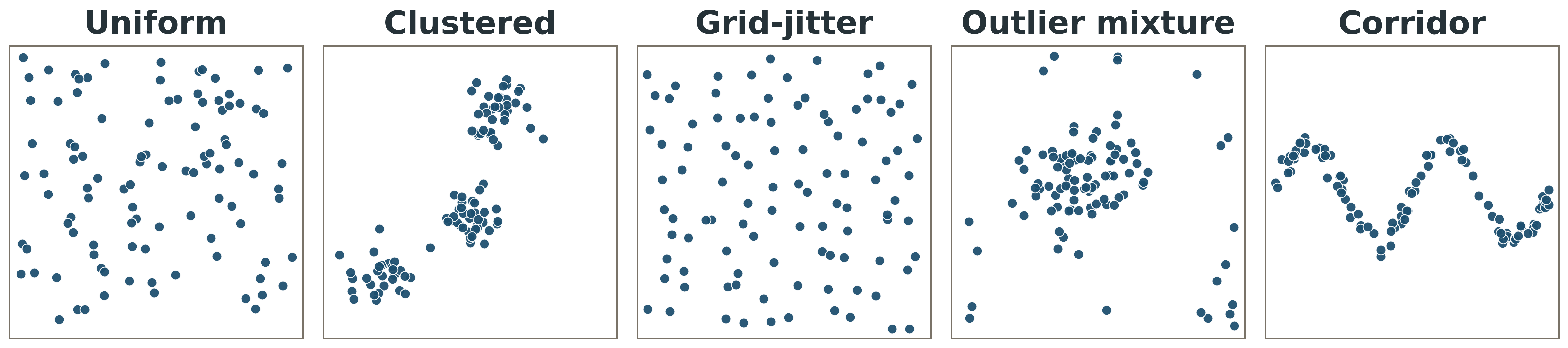}
\caption{The five spatial distributions used in our benchmark (\texttt{EUC\_2D}, $N{=}100$). Each distribution $\times$ distance type combination forms one TSP family.}
\label{fig:distributions}
\end{figure}

\vspace{-1mm}

We test the two-stage method on a benchmark of twenty TSP families: four distance types \cite{reineltTSPLIBTravelingSalesman1991} (\texttt{EUC\_2D}, \texttt{MAN\_2D}, \texttt{ATT}, \texttt{GEO}) crossed with five spatial distributions (Figure~\ref{fig:distributions}). \texttt{EUC\_2D} is rounded Euclidean, \texttt{MAN\_2D} is Manhattan, \texttt{GEO} is great-circle distance on a sphere of radius 6378.388\,km, and \texttt{ATT} is pseudo-Euclidean with ceiling rounding: $d_{ij} = \lceil\sqrt{(\Delta x^2 + \Delta y^2)/10}\,\rceil$. Training uses 5\,000 instances per family at $N{=}100$, giving 100\,000 training instances in total. We train at $N{=}100$ because Concorde labelling cost scales with $O(N^3)$, making this the largest size where 100\,000 labelled instances remain computationally tractable; the transfer to $N{=}50, 200, 500$ then tests whether the learned features generalise across problem scales. Validation uses 1\,000 instances per family at $N{=}100$. Each test split at $N{=}50, 100, 200, 500$ uses 1\,000 instances per family.

We train one model per classifier type (LR, SVM, XGBoost), each trained once on the combined 100,000-instance dataset spanning all 20 families. The main experiments use the union base candidate graph; an ablation study compares all three Stage~1 generators ($\alpha$-Nearest, POPMUSIC, and their union).

\vspace{-3mm}
\subsection{Labels and Parameter Settings}\label{sec:experiment_params}
\vspace{-1mm}
We obtain exact-tour labels with Concorde \cite{applegate2006concorde} on the candidate edges only, following \cite{sunUsingStatisticalMeasures2021,sunGeneralizationMachineLearning2021}. Logistic regression and linear SVM use penalty $C{=}1.0$; XGBoost uses $M{=}96$ rounds, $\texttt{max\_depth}{=}6$, learning rate $\alpha_{\mathrm{lr}}{=}0.08$. For the pruning rule (Section~\ref{sec:pruning}), we fix the temperature $T{=}1$ because $T$ and $\eta$ both control how cumulative softmax mass is allocated and tuning both introduces a redundant degree of freedom. The hard floor $m_{\min}{=}2$ is a structural feasibility requirement (every node needs two incident edges to participate in a Hamiltonian cycle), not a tuned hyperparameter. We select $\eta$ on the validation set: among all values reaching ${\geq}99\%$ coverage, we pick the one yielding the fewest retained edges, which gives $\eta{=}0.60$ for all three models. Pruned graphs go to LKH at its default configuration, with CPU times measured on an AMD EPYC 9634. Each experiment is independently run 10 times with random seeds $\{1,\dots,10\}$. Exact LKH, XGBoost, and Concorde versions, and the instance generators are listed in the anonymised supplementary repository.

\begin{figure}[!h]
\centering
\includegraphics[width=\columnwidth]{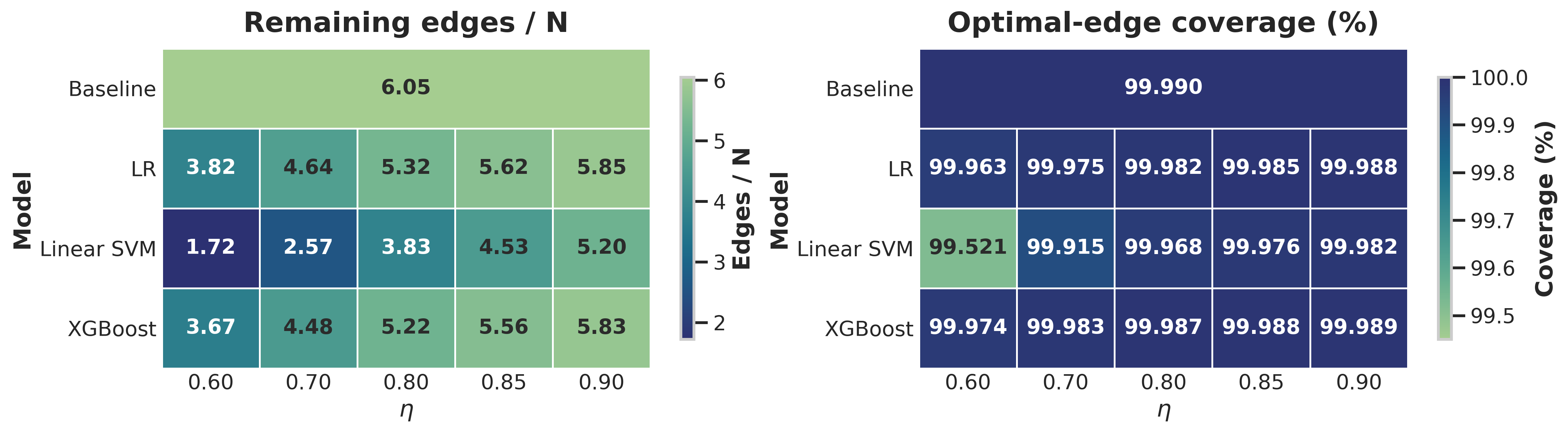}
\vspace{-6mm}
\caption{Sensitivity of the pruning threshold $\eta$ on the union base graph ($N{=}100$, validation set). Left: retained density (Edges/$N$). Right: optimal-tour coverage (\%). }
\label{fig:eta_sensitivity}
\end{figure}

Figure~\ref{fig:eta_sensitivity} shows how the $\eta$ setting affects the density--coverage trade-off. XGBoost and LR are stable: coverage stays above $99.9\%$ for $\eta \geq 0.70$, while SVM is more sensitive, dropping to $99.5\%$ at $\eta{=}0.60$. At $\eta{=}0.50$, SVM coverage falls below $99\%$, so $\eta{=}0.60$ is the most aggressive value at which all three models meet the ${\geq}99\%$-coverage criterion of Section~\ref{sec:pruning}. We use $\eta{=}0.60$ throughout.

\vspace{-3mm}
\section{Results and Analysis}
\vspace{-2mm}
\subsection{Main Results}\label{sec:main}
\vspace{-1mm}

Table~\ref{tab:main_results} presents results of the proposed two-stage method across four problem sizes. A single model is trained at $N{=}100$ and evaluated at out-of-training sizes $50$, $200$, and $500$ with 10 random seeds, so the table doubles as a cross-size transfer test.

The union base graph reaches near-perfect coverage (${\geq}99.96\%$) but is dense at ${\sim}6N$ edges. Pruning removes $37\%$--$47\%$ of these edges while keeping coverage above $99.69\%$ across all sizes. The density reduction improves with scale: XGBoost prunes $37\%$ at TSP50 and $47\%$ at TSP500, while its coverage declines only from $99.987\%$ to $99.693\%$---a $0.29$\,pp drop over a $10\times$ increase in problem size. Relative to the complete graph, the pruned graphs are $6$--$15\times$ sparser at TSP50 ($1.61N$--$4.06N$ vs.\ $24.5N$) and $76$--$114\times$ sparser at TSP500 ($2.19N$--$3.30N$ vs.\ $249.5N$). The pruning step adds at most $83$\,ms per instance (Table~\ref{tab:runtime_main}).

\begin{table*}[!h]
\caption{Two-Stage Pruning Results on the Union Base Graph\label{tab:main_results}}
\vspace{-3mm}
\centering
\scriptsize
\setlength{\tabcolsep}{3.5pt}
\resizebox{\columnwidth}{!}{%
\begin{tabular}{ccccccccc}
\toprule
\multirow{2}{*}{Method}
& \multicolumn{2}{c}{TSP50} & \multicolumn{2}{c}{TSP100} & \multicolumn{2}{c}{TSP200} & \multicolumn{2}{c}{TSP500} \\
\cmidrule(lr){2-3}\cmidrule(lr){4-5}\cmidrule(lr){6-7}\cmidrule(lr){8-9}
& Edges/$N$\,$\downarrow$ & Cov\,(\%)\,$\uparrow$
& Edges/$N$\,$\downarrow$ & Cov\,(\%)\,$\uparrow$
& Edges/$N$\,$\downarrow$ & Cov\,(\%)\,$\uparrow$
& Edges/$N$\,$\downarrow$ & Cov\,(\%)\,$\uparrow$ \\
\midrule
Complete graph
& $24.5$   & $100.000$
& $49.5$   & $100.000$
& $99.5$   & $100.000$
& $249.5$   & $100.000$ \\
\addlinespace[2pt]
Union base
& $6.25$          & $100.000$
& $6.05$          & $99.991$
& $5.92$          & $99.959$
& $5.93$          & $99.966$ \\
& {\color{gray!60}\scriptsize$\pm0.59$} & {\color{gray!60}\scriptsize$\pm0.008$}
& {\color{gray!60}\scriptsize$\pm0.44$} & {\color{gray!60}\scriptsize$\pm0.051$}
& {\color{gray!60}\scriptsize$\pm0.26$} & {\color{gray!60}\scriptsize$\pm0.074$}
& {\color{gray!60}\scriptsize$\pm0.16$} & {\color{gray!60}\scriptsize$\pm0.043$} \\
\addlinespace[2pt]
\;+ LR
& $4.06$          & $99.986$
& $3.84$          & $99.963$
& $3.42$          & $99.746$
& $3.30$          & $99.427$ \\
& {\color{gray!60}\scriptsize$\pm0.41$} & {\color{gray!60}\scriptsize$\pm0.156$}
& {\color{gray!60}\scriptsize$\pm0.33$} & {\color{gray!60}\scriptsize$\pm0.216$}
& {\color{gray!60}\scriptsize$\pm0.19$} & {\color{gray!60}\scriptsize$\pm0.391$}
& {\color{gray!60}\scriptsize$\pm0.11$} & {\color{gray!60}\scriptsize$\pm0.341$} \\
\addlinespace[2pt]
\;+ SVM
& $\mathbf{1.61}$ & $99.834$
& $\mathbf{1.79}$ & $99.701$
& $\mathbf{1.96}$ & $99.109$
& $\mathbf{2.19}$ & $98.601$ \\
& {\color{gray!60}\scriptsize$\pm0.20$} & {\color{gray!60}\scriptsize$\pm0.688$}
& {\color{gray!60}\scriptsize$\pm0.19$} & {\color{gray!60}\scriptsize$\pm0.664$}
& {\color{gray!60}\scriptsize$\pm0.08$} & {\color{gray!60}\scriptsize$\pm0.762$}
& {\color{gray!60}\scriptsize$\pm0.04$} & {\color{gray!60}\scriptsize$\pm0.555$} \\
\addlinespace[2pt]
\;+ XGBoost
& $3.94$          & $\mathbf{99.987}$
& $3.64$          & $\mathbf{99.976}$
& $3.24$          & $\mathbf{99.821}$
& $3.14$          & $\mathbf{99.693}$ \\
& {\color{gray!60}\scriptsize$\pm0.42$} & {\color{gray!60}\scriptsize$\pm0.146$}
& {\color{gray!60}\scriptsize$\pm0.30$} & {\color{gray!60}\scriptsize$\pm0.144$}
& {\color{gray!60}\scriptsize$\pm0.17$} & {\color{gray!60}\scriptsize$\pm0.295$}
& {\color{gray!60}\scriptsize$\pm0.10$} & {\color{gray!60}\scriptsize$\pm0.215$} \\
\bottomrule
\end{tabular}
}
\vspace{0.3em}
\par\noindent{\scriptsize Edges/$N$: retained edges per node. Cov: optimal-tour coverage. All models trained once at $N{=}100$. {\color{gray!60}Standard deviations over 10 independent runs shown in gray.} Wilcoxon rank-sum test vs.\ union base: all coverage differences are statistically significant ($p < 0.05$).}
\end{table*}

\begin{table}[!h]
\caption{LKH Solver Performance on Pruned Union Graphs\label{tab:lkh_main}}
\vspace{-3mm}
\centering
\scriptsize
\setlength{\tabcolsep}{2.5pt}
\resizebox{\columnwidth}{!}{%
\begin{tabular}{ccccccccccccc}
\toprule
\multirow{2}{*}{Method}
& \multicolumn{3}{c}{TSP50} & \multicolumn{3}{c}{TSP100} & \multicolumn{3}{c}{TSP200} & \multicolumn{3}{c}{TSP500} \\
\cmidrule(lr){2-4}\cmidrule(lr){5-7}\cmidrule(lr){8-10}\cmidrule(lr){11-13}
& CPU\,$\downarrow$ & Spd.\,$\uparrow$ & Gap\,$\downarrow$
& CPU\,$\downarrow$ & Spd.\,$\uparrow$ & Gap\,$\downarrow$
& CPU\,$\downarrow$ & Spd.\,$\uparrow$ & Gap\,$\downarrow$
& CPU\,$\downarrow$ & Spd.\,$\uparrow$ & Gap\,$\downarrow$ \\
\midrule
  $\alpha$-Near.\ base
  & $0.032$ & --- & $0.003$
  & $0.129$ & --- & $0.146$
  & $0.335$ & --- & $0.193$
  & $1.618$ & --- & $0.305$ \\
  & {\color{gray!60}\scriptsize$\pm0.019$} & & {\color{gray!60}\scriptsize$\pm0.048$}
  & {\color{gray!60}\scriptsize$\pm0.077$} & & {\color{gray!60}\scriptsize$\pm0.230$}
  & {\color{gray!60}\scriptsize$\pm0.203$} & & {\color{gray!60}\scriptsize$\pm0.254$}
  & {\color{gray!60}\scriptsize$\pm0.798$} & & {\color{gray!60}\scriptsize$\pm0.216$} \\
  \addlinespace[2pt]
  POPMUSIC base
  & $0.029$ & --- & $0.003$
  & $0.132$ & --- & $0.131$
  & $0.413$ & --- & $0.173$
  & $1.784$ & --- & $0.274$ \\
  & {\color{gray!60}\scriptsize$\pm0.018$} & & {\color{gray!60}\scriptsize$\pm0.045$}
  & {\color{gray!60}\scriptsize$\pm0.079$} & & {\color{gray!60}\scriptsize$\pm0.208$}
  & {\color{gray!60}\scriptsize$\pm0.248$} & & {\color{gray!60}\scriptsize$\pm0.229$}
  & {\color{gray!60}\scriptsize$\pm0.879$} & & {\color{gray!60}\scriptsize$\pm0.196$} \\
\midrule
Union base
& $0.030$ & ---     & $0.003$
& $0.125$ & ---     & $0.138$
& $0.408$ & ---     & $0.182$
& $1.670$ & ---     & $0.289$ \\
& {\color{gray!60}\scriptsize$\pm0.018$} &                          & {\color{gray!60}\scriptsize$\pm0.046$}
& {\color{gray!60}\scriptsize$\pm0.074$} &                          & {\color{gray!60}\scriptsize$\pm0.218$}
& {\color{gray!60}\scriptsize$\pm0.244$} &                          & {\color{gray!60}\scriptsize$\pm0.239$}
& {\color{gray!60}\scriptsize$\pm0.824$} &                          & {\color{gray!60}\scriptsize$\pm0.205$} \\
\addlinespace[2pt]
\;+ LR
& $\mathbf{0.024}$ & $(\mathbf{1.28})\times$ & $0.003$
& $\mathbf{0.104}$ & $(\mathbf{1.21})\times$ & $\mathbf{0.124}$
& $\mathbf{0.336}$ & $(\mathbf{1.22})\times$ & $\mathbf{0.164}$
& $\mathbf{1.399}$ & $(\mathbf{1.20})\times$ & $\mathbf{0.261}$ \\
& {\color{gray!60}\scriptsize$\pm0.016$} & {\color{gray!60}\scriptsize$\pm0.17$}  & {\color{gray!60}\scriptsize$\pm0.043$}
& {\color{gray!60}\scriptsize$\pm0.061$} & {\color{gray!60}\scriptsize$\pm0.07$}  & {\color{gray!60}\scriptsize$\pm0.199$}
& {\color{gray!60}\scriptsize$\pm0.201$} & {\color{gray!60}\scriptsize$\pm0.06$}  & {\color{gray!60}\scriptsize$\pm0.220$}
& {\color{gray!60}\scriptsize$\pm0.687$} & {\color{gray!60}\scriptsize$\pm0.07$}  & {\color{gray!60}\scriptsize$\pm0.189$} \\
\addlinespace[2pt]
\;+ SVM
& $0.028$ & $(1.09)\times$ & $0.003$
& $0.142$ & $(0.89)\times$ & $0.130$
& $0.374$ & $(1.09)\times$ & $0.172$
& $1.583$ & $(1.06)\times$ & $0.274$ \\
& {\color{gray!60}\scriptsize$\pm0.017$} & {\color{gray!60}\scriptsize$\pm0.11$}  & {\color{gray!60}\scriptsize$\pm0.044$}
& {\color{gray!60}\scriptsize$\pm0.093$} & {\color{gray!60}\scriptsize$\pm0.05$}  & {\color{gray!60}\scriptsize$\pm0.207$}
& {\color{gray!60}\scriptsize$\pm0.221$} & {\color{gray!60}\scriptsize$\pm0.07$}  & {\color{gray!60}\scriptsize$\pm0.227$}
& {\color{gray!60}\scriptsize$\pm0.772$} & {\color{gray!60}\scriptsize$\pm0.07$}  & {\color{gray!60}\scriptsize$\pm0.195$} \\
\addlinespace[2pt]
\;+ XGBoost
& $0.028$ & $(1.09)\times$ & $\mathbf{0.002}$
& $0.105$ & $(1.20)\times$ & $0.130$
& $0.400$ & $(1.03)\times$ & $0.171$
& $1.593$ & $(1.07)\times$ & $0.274$ \\
& {\color{gray!60}\scriptsize$\pm0.017$} & {\color{gray!60}\scriptsize$\pm0.12$}  & {\color{gray!60}\scriptsize$\pm0.042$}
& {\color{gray!60}\scriptsize$\pm0.061$} & {\color{gray!60}\scriptsize$\pm0.07$}  & {\color{gray!60}\scriptsize$\pm0.201$}
& {\color{gray!60}\scriptsize$\pm0.243$} & {\color{gray!60}\scriptsize$\pm0.05$}  & {\color{gray!60}\scriptsize$\pm0.221$}
& {\color{gray!60}\scriptsize$\pm0.788$} & {\color{gray!60}\scriptsize$\pm0.09$}  & {\color{gray!60}\scriptsize$\pm0.195$} \\
\bottomrule
\end{tabular}
}
\vspace{0.3em}
\par\noindent{\scriptsize CPU: mean LKH solving time (seconds). Spd.: speedup vs.\ unpruned union base. Gap: optimality gap (\%). {\color{gray!60}Standard deviations shown in gray.} Wilcoxon rank-sum test vs.\ union base: all gap differences are statistically significant ($p < 0.05$) except at TSP50.}
\end{table}

Across models, LR and XGBoost achieve similar density reductions ($35$--$47\%$), with XGBoost slightly better on coverage at larger sizes ($99.69\%$ vs.\ $99.43\%$ at TSP500). SVM prunes the hardest: it retains the fewest edges at every size (e.g., $1.96N$ at TSP200, versus $3.24N$ for XGBoost) but its coverage is $1.1$--$1.6$\,pp lower than the next-best model at TSP100--500, dropping to $98.60\%$ at TSP500. That linear logistic regression suffices reflects the structure of the problem that source-provenance creates: as the failure-mode analysis in Section~\ref{sec:features_failures} shows, dual-source edges are almost always optimal, so a linear decision boundary on the single-source subset is sufficient.

When the pruned graphs are passed to LKH without re-tuning, the sparser candidate graph yields faster search at unchanged solution quality (Table~\ref{tab:lkh_main}). Union + LR delivers $1.21$--$1.28\times$ speedup at every size; XGBoost gives $1.20\times$ at TSP100 but only $1.03$--$1.09\times$ elsewhere; SVM slows TSP100 to $0.89\times$ because its sharper pruning removes edges LKH still uses as search guidance. The optimality gap never increases: every pruned model slightly \emph{reduces} the gap relative to the unpruned base (e.g.\ $0.138\% \to 0.124\%$ for LR at TSP100), so the removed edges were redundant for tour quality.

Density reduction and solver speedup remain loosely coupled. At TSP500, LR removes $44\%$ of edges and yields $1.20\times$ speedup, while XGBoost removes $47\%$ but only $1.07\times$. LKH treats candidate edges as search guidance, not just constraints, so statistically redundant edges can still affect LKH move choices; LR's lighter pruning preserves more of this connectivity.

\vspace{-3mm}
\subsection{Comparison with Neural Sparsification Methods}\label{sec:neural_comparison}
\vspace{-1mm}

We compare our two-stage method against three neural approaches that produce edge heatmaps usable as implicit sparsifiers: DIFUSCO~\cite{sunDIFUSCOGraphbasedDiffusion2024}, a diffusion-based model; AttGCN~\cite{fuGeneralizeSmallPretrained2021}, an attention-based graph convolutional network; and DIMES~\cite{qiuDIMESDifferentiableMeta2022}, a differentiable meta-solver. We note that DIFUSCO and DIMES are primarily end-to-end solvers; we evaluate them here specifically on the quality of their implicit edge selection. These neural methods take city coordinates as input and produce edge heatmaps over the complete graph; they are trained exclusively on Euclidean instances. To ensure a fair comparison, we restrict the evaluation to \texttt{EUC\_2D} TSP500 across all five spatial distributions.

\begin{table*}[!b]
\centering
\scriptsize
\setlength{\tabcolsep}{3.2pt}
\renewcommand{\arraystretch}{1.05}
\caption{Comparison with Neural Sparsification Methods (\texttt{EUC\_2D}, TSP500)\label{tab:tsp500_distribution_union_neural}}
\vspace{-3mm}
\resizebox{\textwidth}{!}{%
\begin{tabular}{lcccccccccc}
\toprule
\multirow{2}{*}{Algorithm}
& \multicolumn{2}{c}{uniform} & \multicolumn{2}{c}{clustered} & \multicolumn{2}{c}{grid\_jitter} & \multicolumn{2}{c}{outlier\_mixture} & \multicolumn{2}{c}{corridor} \\
\cmidrule(lr){2-3}\cmidrule(lr){4-5}\cmidrule(lr){6-7}\cmidrule(lr){8-9}\cmidrule(lr){10-11}
& $E/N$\,$\downarrow$ & Cov\,(\%)\,$\uparrow$
& $E/N$\,$\downarrow$ & Cov\,(\%)\,$\uparrow$
& $E/N$\,$\downarrow$ & Cov\,(\%)\,$\uparrow$
& $E/N$\,$\downarrow$ & Cov\,(\%)\,$\uparrow$
& $E/N$\,$\downarrow$ & Cov\,(\%)\,$\uparrow$ \\
\midrule
DIFUSCO \cite{sunDIFUSCOGraphbasedDiffusion2024}
& $4.513$ & $99.46$
& $4.590$ & $98.19$
& $4.509$ & $99.69$
& $4.673$ & $98.22$
& $4.565$ & $96.85$ \\
& {\color{gray!60}\scriptsize$\pm0.005$} & {\color{gray!60}\scriptsize$\pm0.34$}
& {\color{gray!60}\scriptsize$\pm0.016$} & {\color{gray!60}\scriptsize$\pm0.67$}
& {\color{gray!60}\scriptsize$\pm0.004$} & {\color{gray!60}\scriptsize$\pm0.25$}
& {\color{gray!60}\scriptsize$\pm0.018$} & {\color{gray!60}\scriptsize$\pm0.63$}
& {\color{gray!60}\scriptsize$\pm0.011$} & {\color{gray!60}\scriptsize$\pm0.73$} \\
\addlinespace[2pt]
AttGCN \cite{fuGeneralizeSmallPretrained2021}
& $4.775$ & $99.19$
& $4.775$ & $99.12$
& $4.842$ & $99.26$
& $4.782$ & $99.06$
& $4.338$ & $98.12$ \\
& {\color{gray!60}\scriptsize$\pm0.105$} & {\color{gray!60}\scriptsize$\pm0.41$}
& {\color{gray!60}\scriptsize$\pm0.120$} & {\color{gray!60}\scriptsize$\pm0.41$}
& {\color{gray!60}\scriptsize$\pm0.099$} & {\color{gray!60}\scriptsize$\pm0.38$}
& {\color{gray!60}\scriptsize$\pm0.118$} & {\color{gray!60}\scriptsize$\pm0.46$}
& {\color{gray!60}\scriptsize$\pm0.097$} & {\color{gray!60}\scriptsize$\pm0.68$} \\
\addlinespace[2pt]
DIMES \cite{qiuDIMESDifferentiableMeta2022}
& $3.691$ & $96.35$
& $3.744$ & $94.45$
& $3.725$ & $96.72$
& $4.041$ & $93.44$
& $3.820$ & $87.04$ \\
& {\color{gray!60}\scriptsize$\pm0.038$} & {\color{gray!60}\scriptsize$\pm0.82$}
& {\color{gray!60}\scriptsize$\pm0.131$} & {\color{gray!60}\scriptsize$\pm1.86$}
& {\color{gray!60}\scriptsize$\pm0.027$} & {\color{gray!60}\scriptsize$\pm0.69$}
& {\color{gray!60}\scriptsize$\pm0.040$} & {\color{gray!60}\scriptsize$\pm1.09$}
& {\color{gray!60}\scriptsize$\pm0.040$} & {\color{gray!60}\scriptsize$\pm1.98$} \\
\midrule
Union
& $5.156$ & $100.00$
& $6.141$ & $100.00$
& $5.077$ & $100.00$
& $7.575$ & $100.00$
& $5.691$ & $99.83$ \\
& {\color{gray!60}\scriptsize$\pm0.102$} & {\color{gray!60}\scriptsize$\pm0.01$}
& {\color{gray!60}\scriptsize$\pm0.245$} & {\color{gray!60}\scriptsize$\pm0.01$}
& {\color{gray!60}\scriptsize$\pm0.089$} & {\color{gray!60}\scriptsize$\pm0.00$}
& {\color{gray!60}\scriptsize$\pm0.224$} & {\color{gray!60}\scriptsize$\pm0.02$}
& {\color{gray!60}\scriptsize$\pm0.148$} & {\color{gray!60}\scriptsize$\pm0.18$} \\
\addlinespace[2pt]
Union + LR
& $2.796$ & $99.52$
& $3.408$ & $99.72$
& $2.737$ & $99.56$
& $4.354$ & $99.84$
& $3.210$ & $98.50$ \\
& {\color{gray!60}\scriptsize$\pm0.060$} & {\color{gray!60}\scriptsize$\pm0.31$}
& {\color{gray!60}\scriptsize$\pm0.175$} & {\color{gray!60}\scriptsize$\pm0.25$}
& {\color{gray!60}\scriptsize$\pm0.049$} & {\color{gray!60}\scriptsize$\pm0.29$}
& {\color{gray!60}\scriptsize$\pm0.173$} & {\color{gray!60}\scriptsize$\pm0.18$}
& {\color{gray!60}\scriptsize$\pm0.105$} & {\color{gray!60}\scriptsize$\pm0.68$} \\
\addlinespace[2pt]
Union + SVM
& $2.154$ & $98.61$
& $2.198$ & $98.98$
& $2.164$ & $98.49$
& $2.499$ & $99.42$
& $1.922$ & $97.51$ \\
& {\color{gray!60}\scriptsize$\pm0.034$} & {\color{gray!60}\scriptsize$\pm0.50$}
& {\color{gray!60}\scriptsize$\pm0.058$} & {\color{gray!60}\scriptsize$\pm0.45$}
& {\color{gray!60}\scriptsize$\pm0.031$} & {\color{gray!60}\scriptsize$\pm0.52$}
& {\color{gray!60}\scriptsize$\pm0.051$} & {\color{gray!60}\scriptsize$\pm0.34$}
& {\color{gray!60}\scriptsize$\pm0.049$} & {\color{gray!60}\scriptsize$\pm0.97$} \\
\addlinespace[2pt]
Union + XGBoost
& $2.709$ & $99.91$
& $3.228$ & $99.92$
& $2.656$ & $99.91$
& $4.015$ & $99.95$
& $3.089$ & $98.78$ \\
& {\color{gray!60}\scriptsize$\pm0.056$} & {\color{gray!60}\scriptsize$\pm0.13$}
& {\color{gray!60}\scriptsize$\pm0.144$} & {\color{gray!60}\scriptsize$\pm0.13$}
& {\color{gray!60}\scriptsize$\pm0.046$} & {\color{gray!60}\scriptsize$\pm0.14$}
& {\color{gray!60}\scriptsize$\pm0.142$} & {\color{gray!60}\scriptsize$\pm0.10$}
& {\color{gray!60}\scriptsize$\pm0.096$} & {\color{gray!60}\scriptsize$\pm0.57$} \\
\bottomrule
\end{tabular}%
}
\vspace{0.3em}
\par\noindent{\scriptsize {\color{gray!60}Standard deviations shown in gray.} Wilcoxon rank-sum test: all pairwise coverage differences between Union + ML methods and neural baselines are statistically significant ($p < 0.05$).}
\end{table*}

\begin{table}[!b]
\caption{Per-Instance Time: Neural Methods vs.\ Two-Stage Pipeline\label{tab:runtime_main}}
\vspace{-3mm}
\centering
\scriptsize
\setlength{\tabcolsep}{4pt}
\begin{tabular}{ccccc}
\toprule
Component & TSP50\,$\downarrow$ & TSP100\,$\downarrow$ & TSP200\,$\downarrow$ & TSP500\,$\downarrow$ \\
\midrule
DIFUSCO & 0.332\,s & 0.657\,s & 7.118\,s & 3.880\,s \\
AttGCN & 0.067\,s & 0.104\,s & 0.236\,s & 0.694\,s \\
DIMES & 0.027\,s & 0.038\,s & 0.062\,s & 0.156\,s \\
\midrule
$\alpha$-Nearest prep. & 0.197\,s & 0.642\,s & 2.289\,s & 14.856\,s \\
POPMUSIC prep. & 0.075\,s & 0.186\,s & 0.363\,s & 1.080\,s \\
\midrule
Union + LR prune & 5.93\,ms & 10.56\,ms & 19.67\,ms & 47.64\,ms \\
Union + SVM prune & 5.69\,ms & 10.24\,ms & 19.29\,ms & 46.97\,ms \\
Union + XGB prune & 38.00\,ms & 42.77\,ms & 52.56\,ms & 82.74\,ms \\
\bottomrule
\end{tabular}
\par\noindent{\scriptsize Neural baselines report per-instance wall-clock inference time on a single RTX A5000 GPU (not convertible to CPU time); tested on \texttt{EUC\_2D} only due to model limitations, so times are for reference. DIFUSCO's higher time at TSP200 is not an error: the grid-jitter distribution is disproportionately slow for this model, pulling up the average. Our pipeline reports CPU time on AMD EPYC 9634.}
\end{table}

In coverage, Union + XGBoost obtains the highest value on four of the five distributions ($99.91$--$99.95\%$; Table~\ref{tab:tsp500_distribution_union_neural}); DIFUSCO and AttGCN degrade on harder distributions (DIFUSCO: $96.85\%$ on corridor; DIMES: $87.04\%$); Union + LR matches or exceeds all three neural methods on every distribution. In density, the neural methods operate at $3.7$--$4.8N$ edges, while Union + XGBoost reaches $2.7$--$4.0N$ (sparser on every distribution) and Union + SVM reaches $1.9$--$2.5N$. The neural baselines cover only \texttt{EUC\_2D}.

In runtime, Stage~2 pruning adds under $83$\,ms per instance on CPU (Table~\ref{tab:runtime_main}). The pipeline is dominated by Stage~1: $\alpha$-Nearest ($0.20$--$14.9$\,s) and POPMUSIC ($0.08$--$1.08$\,s). At TSP500 Stage~1 is $200$--$340\times$ slower than Stage~2, so any further wall-clock improvement has to come from Stage~1 (e.g.\ parallelisation or candidate caching), not from the learned-pruning step. The neural baselines require GPU, with DIFUSCO taking up to $7.1$\,s per instance.

\subsection{Ablation: Base Graph Choice}\label{sec:ablation}
\vspace{-1mm}

\begin{table}[!h]
\caption{Ablation: Effect of Base Graph Choice (XGBoost)\label{tab:ablation}}
\vspace{-3mm}
\centering
\scriptsize
\setlength{\tabcolsep}{1.5pt}
\resizebox{\columnwidth}{!}{%
\begin{tabular}{ccccccccccccc}
\toprule
\multirow{2}{*}{Method}
& \multicolumn{3}{c}{TSP50} & \multicolumn{3}{c}{TSP100} & \multicolumn{3}{c}{TSP200} & \multicolumn{3}{c}{TSP500} \\
\cmidrule(lr){2-4}\cmidrule(lr){5-7}\cmidrule(lr){8-10}\cmidrule(lr){11-13}
& E/$N$\,$\downarrow$ & Cov\,$\uparrow$ & Spd.\,$\uparrow$
& E/$N$\,$\downarrow$ & Cov\,$\uparrow$ & Spd.\,$\uparrow$
& E/$N$\,$\downarrow$ & Cov\,$\uparrow$ & Spd.\,$\uparrow$
& E/$N$\,$\downarrow$ & Cov\,$\uparrow$ & Spd.\,$\uparrow$ \\
\midrule
Full graph + XGB
& ---     & ---       & ---
& $31.65$ & $99.996$  & ---
& ---     & ---       & ---
& ---     & ---       & ---     \\
&         &                         &
& {\color{gray!60}\scriptsize$\pm0.35$} & {\color{gray!60}\scriptsize$\pm0.039$} &
&         &                         &
&         &                         &         \\
\midrule
$\alpha$-Near.\ base
& $6.25$ & $99.984$ & ---
& $6.03$ & $99.718$ & ---
& $5.80$ & $99.756$ & ---
& $5.74$ & $99.918$ & ---   \\
& {\color{gray!60}\scriptsize$\pm0.59$} & {\color{gray!60}\scriptsize$\pm0.142$} &
& {\color{gray!60}\scriptsize$\pm0.45$} & {\color{gray!60}\scriptsize$\pm0.242$} &
& {\color{gray!60}\scriptsize$\pm0.27$} & {\color{gray!60}\scriptsize$\pm0.148$} &
& {\color{gray!60}\scriptsize$\pm0.17$} & {\color{gray!60}\scriptsize$\pm0.079$} &   \\
\addlinespace[2pt]
$\alpha$-Near.\ + XGB
& $4.89$ & $99.158$ & $(\mathbf{1.12})\times$
& $4.42$ & $99.367$ & $(1.13)\times$
& $4.12$ & $99.179$ & $(0.90)\times$
& $4.60$ & $98.601$ & $(1.06)\times$ \\
& {\color{gray!60}\scriptsize$\pm0.46$} & {\color{gray!60}\scriptsize$\pm1.173$} & {\color{gray!60}\scriptsize$\pm0.04$}
& {\color{gray!60}\scriptsize$\pm0.34$} & {\color{gray!60}\scriptsize$\pm0.643$} & {\color{gray!60}\scriptsize$\pm0.05$}
& {\color{gray!60}\scriptsize$\pm0.17$} & {\color{gray!60}\scriptsize$\pm0.519$} & {\color{gray!60}\scriptsize$\pm0.04$}
& {\color{gray!60}\scriptsize$\pm0.11$} & {\color{gray!60}\scriptsize$\pm0.526$} & {\color{gray!60}\scriptsize$\pm0.07$} \\
\midrule
POPMUSIC base
& $1.14$ & $99.952$ & ---
& $1.29$ & $99.918$ & ---
& $1.96$ & $99.717$ & ---
& $2.37$ & $99.557$ & ---   \\
& {\color{gray!60}\scriptsize$\pm0.14$} & {\color{gray!60}\scriptsize$\pm0.548$} &
& {\color{gray!60}\scriptsize$\pm0.14$} & {\color{gray!60}\scriptsize$\pm0.587$} &
& {\color{gray!60}\scriptsize$\pm0.09$} & {\color{gray!60}\scriptsize$\pm0.502$} &
& {\color{gray!60}\scriptsize$\pm0.05$} & {\color{gray!60}\scriptsize$\pm0.279$} &   \\
\addlinespace[2pt]
POPMUSIC + XGB
& $\mathbf{1.13}$ & $99.913$ & $(1.03)\times$
& $\mathbf{1.28}$ & $99.763$ & $(1.14)\times$
& $\mathbf{1.85}$ & $99.197$ & $(\mathbf{1.08})\times$
& $\mathbf{2.22}$ & $99.187$ & $(\mathbf{1.24})\times$ \\
& {\color{gray!60}\scriptsize$\pm0.14$} & {\color{gray!60}\scriptsize$\pm0.661$} & {\color{gray!60}\scriptsize$\pm0.04$}
& {\color{gray!60}\scriptsize$\pm0.13$} & {\color{gray!60}\scriptsize$\pm0.758$} & {\color{gray!60}\scriptsize$\pm0.03$}
& {\color{gray!60}\scriptsize$\pm0.08$} & {\color{gray!60}\scriptsize$\pm0.785$} & {\color{gray!60}\scriptsize$\pm0.08$}
& {\color{gray!60}\scriptsize$\pm0.05$} & {\color{gray!60}\scriptsize$\pm0.435$} & {\color{gray!60}\scriptsize$\pm0.10$} \\
\midrule
Union base
& $6.25$ & $100.000$ & ---
& $6.05$ & $99.991$  & ---
& $5.92$ & $99.959$  & ---
& $5.93$ & $99.966$  & ---   \\
& {\color{gray!60}\scriptsize$\pm0.59$} & {\color{gray!60}\scriptsize$\pm0.008$} &
& {\color{gray!60}\scriptsize$\pm0.44$} & {\color{gray!60}\scriptsize$\pm0.050$} &
& {\color{gray!60}\scriptsize$\pm0.26$} & {\color{gray!60}\scriptsize$\pm0.074$} &
& {\color{gray!60}\scriptsize$\pm0.16$} & {\color{gray!60}\scriptsize$\pm0.043$} &   \\
\addlinespace[2pt]
Union + XGB
& $3.94$ & $\mathbf{99.987}$ & $(1.09)\times$
& $3.64$ & $\mathbf{99.976}$ & $(\mathbf{1.20})\times$
& $3.24$ & $\mathbf{99.821}$ & $(1.03)\times$
& $3.14$ & $\mathbf{99.693}$ & $(1.07)\times$ \\
& {\color{gray!60}\scriptsize$\pm0.42$} & {\color{gray!60}\scriptsize$\pm0.146$} & {\color{gray!60}\scriptsize$\pm0.12$}
& {\color{gray!60}\scriptsize$\pm0.30$} & {\color{gray!60}\scriptsize$\pm0.144$} & {\color{gray!60}\scriptsize$\pm0.07$}
& {\color{gray!60}\scriptsize$\pm0.17$} & {\color{gray!60}\scriptsize$\pm0.295$} & {\color{gray!60}\scriptsize$\pm0.05$}
& {\color{gray!60}\scriptsize$\pm0.10$} & {\color{gray!60}\scriptsize$\pm0.215$} & {\color{gray!60}\scriptsize$\pm0.09$} \\
\bottomrule
\end{tabular}
}
\vspace{0.3em}
\par\noindent{\scriptsize E/$N$: retained edges per node. Cov: optimal-tour coverage (\%). Spd.: LKH speedup vs.\ unpruned base of the same candidate. {\color{gray!60}Standard deviations shown in gray.}}
\end{table}

As a full-graph pruning baseline (Table~\ref{tab:ablation}), applying XGBoost directly to the complete graph at TSP100 yields $31.65N$ edges with $99.996\%$ coverage; Union + XGBoost reaches almost the same coverage ($99.976\%$) at $3.64N$, an $8.7\times$ density reduction. Without the Stage~1 candidate restriction and provenance indicators, the model must score $N(N{-}1)/2$ edges with no consensus cue and a $1{:}N/2$ positive-to-negative ratio, which makes the learning problem substantially harder.

Union + XGBoost reduces density by $37$--$47\%$ at every size while retaining ${\geq}99.693\%$ coverage. On $\alpha$-Nearest, XGBoost still cuts density but the pruning model trained at $N{=}100$ does not transfer well to $N{=}500$: coverage drops to $98.601\%$, down from a base coverage of $99.918\%$. On POPMUSIC, XGBoost degrades coverage: the base is already very sparse at $1.14$--$2.37N$, with few removable non-tour candidate edges, so pruning reduces coverage at every size.

From TSP200 onward, union + XGBoost surpasses POPMUSIC base in coverage ($99.821\%$ vs.\ $99.717\%$ at TSP200; $99.693\%$ vs.\ $99.557\%$ at TSP500). POPMUSIC dominates among single-stage heuristics at TSP50 ($99.952\%$) but degrades with scale.

\vspace{-3mm}
\subsection{Performance Across Distance Types and Distributions}\label{sec:per_family}
\vspace{-1mm}

\begin{table}[!h]
\caption{Per-Distance-Type Breakdown (Union + XGBoost)\label{tab:per_distance}}
\vspace{-3mm}
\centering
\scriptsize
\setlength{\tabcolsep}{3pt}
\resizebox{\columnwidth}{!}{%
\begin{tabular}{ccccccccc}
\toprule
 & \multicolumn{2}{c}{TSP50} & \multicolumn{2}{c}{TSP100} & \multicolumn{2}{c}{TSP200} & \multicolumn{2}{c}{TSP500} \\
\cmidrule(lr){2-3}\cmidrule(lr){4-5}\cmidrule(lr){6-7}\cmidrule(lr){8-9}
Distance & Edges/$N$\,$\downarrow$ & Cov\,(\%)\,$\uparrow$ & Edges/$N$\,$\downarrow$ & Cov\,(\%)\,$\uparrow$ & Edges/$N$\,$\downarrow$ & Cov\,(\%)\,$\uparrow$ & Edges/$N$\,$\downarrow$ & Cov\,(\%)\,$\uparrow$ \\
\midrule
ATT & 4.16 & 99.992 & 3.74 & 99.974 & 3.29 & 99.775 & 3.17 & 99.580 \\
EUC\_2D & 4.15 & 99.989 & 3.74 & 99.978 & 3.29 & 99.787 & 3.15 & 99.595 \\
GEO & 3.60 & 99.986 & 3.47 & 99.984 & 3.16 & 99.898 & 3.09 & 99.834 \\
MAN\_2D & 3.85 & 99.981 & 3.59 & 99.969 & 3.23 & 99.824 & 3.15 & 99.763 \\
\bottomrule
\end{tabular}
}
\vspace{0.3em}
\end{table}

\begin{table}[!h]
\caption{Per-Distribution Breakdown (Union + XGBoost)\label{tab:per_distribution}}
\vspace{-3mm}
\centering
\scriptsize
\setlength{\tabcolsep}{3pt}
\resizebox{\columnwidth}{!}{%
\begin{tabular}{ccccccccc}
\toprule
 & \multicolumn{2}{c}{TSP50} & \multicolumn{2}{c}{TSP100} & \multicolumn{2}{c}{TSP200} & \multicolumn{2}{c}{TSP500} \\
\cmidrule(lr){2-3}\cmidrule(lr){4-5}\cmidrule(lr){6-7}\cmidrule(lr){8-9}
Distribution & Edges/$N$\,$\downarrow$ & Cov\,(\%)\,$\uparrow$ & Edges/$N$\,$\downarrow$ & Cov\,(\%)\,$\uparrow$ & Edges/$N$\,$\downarrow$ & Cov\,(\%)\,$\uparrow$ & Edges/$N$\,$\downarrow$ & Cov\,(\%)\,$\uparrow$ \\
\midrule
Uniform & 3.09 & 100.000 & 2.88 & 99.997 & 2.66 & 99.942 & 2.68 & 99.854 \\
Clustered & 4.10 & 99.985 & 3.80 & 99.992 & 3.41 & 99.942 & 3.27 & 99.913 \\
Grid-jitter & 2.97 & 99.999 & 2.65 & 99.999 & 2.57 & 99.940 & 2.63 & 99.859 \\
Outlier-mix & 5.95 & 99.998 & 5.26 & 99.993 & 4.46 & 99.968 & 4.03 & 99.906 \\
Corridor & 4.44 & 99.976 & 3.59 & 99.945 & 3.15 & 99.317 & 3.14 & 98.960 \\
\bottomrule
\end{tabular}
}
\vspace{0.3em}
\end{table}

Across distance types, coverage varies by less than $0.3$ percentage points across all four types at every size (Table~\ref{tab:per_distance}): at TSP50 the range is $99.981$--$99.992\%$, and at TSP500 $99.580$--$99.834\%$. Density also decreases uniformly with $N$ (from ${\sim}4N$ at TSP50 to ${\sim}3N$ at TSP500). The same distance-type-agnostic model handles all four types without per-type tuning.

Across distributions, four of the five retain ${\geq}99.85\%$ coverage at every size (Table~\ref{tab:per_distribution}). Grid-jitter and uniform are the sparsest at $2.6$--$3.1N$ and stay near-perfect; outlier-mixture needs $4$--$6N$ density because of its irregular local neighbourhoods but still keeps $99.91\%$ at TSP500.

Corridor is the hard case. Coverage drops from $99.976\%$ at TSP50 to $98.960\%$ at TSP500, with the largest drop occurring between TSP100 ($99.945\%$) and TSP200 ($99.317\%$). Density decreases from $4.44N$ at TSP50 to $3.14N$ at TSP500.

Table~\ref{tab:ablation_corridor} focuses on ATT corridor, the hardest family. The failure separates into two layers that act on different edges. \textbf{Layer 1 (Stage~1 recall)}: $\alpha$-Nearest misses on average $1.35$ optimal edges per instance and POPMUSIC misses $3.40$, but on largely disjoint edge sets ($\alpha$-Nearest fails on geometric extremes that lie outside its five-nearest-neighbour reach, whereas POPMUSIC fails on moderate-length edges crossing its subproblem boundaries), so their union misses only $0.40$ edges per instance on average. \textbf{Layer 2 (Stage~2 pruning)}: the residual loss is what the learned model prunes from the base---short, locally atypical edges (high edge rank, non-mutual neighbours), as the failure-mode analysis in Section~\ref{sec:features_failures} shows. Stage~1 repairs the recall failures, and Stage~2 decides which single-source edges to keep.

\begin{table}[!h]
\caption{Hard Case: ATT Corridor (XGBoost)\label{tab:ablation_corridor}}
\vspace{-3mm}
\centering
\scriptsize
\setlength{\tabcolsep}{3pt}
\resizebox{\columnwidth}{!}{%
\begin{tabular}{ccccccccc}
\toprule
\multirow{2}{*}{Method}
& \multicolumn{4}{c}{TSP100} & \multicolumn{4}{c}{TSP500} \\
\cmidrule(lr){2-5}\cmidrule(lr){6-9}
& Edges/$N$\,$\downarrow$ & Cov\,(\%)\,$\uparrow$ & Spd.\,$\uparrow$ & Gap\,(\%)\,$\downarrow$
& Edges/$N$\,$\downarrow$ & Cov\,(\%)\,$\uparrow$ & Spd.\,$\uparrow$ & Gap\,(\%)\,$\downarrow$ \\
\midrule
$\alpha$-Near.\ base
& $5.73$ & $98.348$ & ---            & $0.146$
& $5.50$ & $99.637$ & ---            & $0.305$ \\
& {\color{gray!60}\scriptsize$\pm0.78$} & {\color{gray!60}\scriptsize$\pm0.673$} &                          & {\color{gray!60}\scriptsize$\pm0.288$}
& {\color{gray!60}\scriptsize$\pm0.15$} & {\color{gray!60}\scriptsize$\pm0.264$} &                          & {\color{gray!60}\scriptsize$\pm0.605$} \\
\addlinespace[2pt]
$\alpha$-Near.\ + XGB
& $4.20$ & $98.007$ & $(1.04)\times$ & $0.136$
& $4.53$ & $97.308$ & $(1.06)\times$ & $0.289$ \\
& {\color{gray!60}\scriptsize$\pm0.61$} & {\color{gray!60}\scriptsize$\pm0.832$} & {\color{gray!60}\scriptsize$\pm0.07$}   & {\color{gray!60}\scriptsize$\pm0.282$}
& {\color{gray!60}\scriptsize$\pm0.10$} & {\color{gray!60}\scriptsize$\pm0.732$} & {\color{gray!60}\scriptsize$\pm0.07$}   & {\color{gray!60}\scriptsize$\pm0.592$} \\
\midrule
POPMUSIC base
& $1.28$ & $99.677$ & ---            & $0.131$
& $1.91$ & $96.696$ & ---            & $0.274$ \\
& {\color{gray!60}\scriptsize$\pm0.11$} & {\color{gray!60}\scriptsize$\pm1.369$} &                          & {\color{gray!60}\scriptsize$\pm0.275$}
& {\color{gray!60}\scriptsize$\pm0.06$} & {\color{gray!60}\scriptsize$\pm1.444$} &                          & {\color{gray!60}\scriptsize$\pm0.578$} \\
\addlinespace[2pt]
POPMUSIC + XGB
& $1.26$ & $99.609$ & $(1.10)\times$ & $0.137$
& $1.83$ & $96.449$ & $(1.13)\times$ & $0.289$ \\
& {\color{gray!60}\scriptsize$\pm0.11$} & {\color{gray!60}\scriptsize$\pm1.388$} & {\color{gray!60}\scriptsize$\pm0.10$}   & {\color{gray!60}\scriptsize$\pm0.281$}
& {\color{gray!60}\scriptsize$\pm0.05$} & {\color{gray!60}\scriptsize$\pm1.443$} & {\color{gray!60}\scriptsize$\pm0.15$}   & {\color{gray!60}\scriptsize$\pm0.591$} \\
\midrule
Union base
& $5.82$ & $99.945$ & ---            & $0.138$
& $5.71$ & $99.781$ & ---            & $0.289$ \\
& {\color{gray!60}\scriptsize$\pm0.76$} & {\color{gray!60}\scriptsize$\pm0.268$} &                          & {\color{gray!60}\scriptsize$\pm0.281$}
& {\color{gray!60}\scriptsize$\pm0.14$} & {\color{gray!60}\scriptsize$\pm0.211$} &                          & {\color{gray!60}\scriptsize$\pm0.591$} \\
\addlinespace[2pt]
Union + XGB
& $3.55$ & $99.898$ & $(1.29)\times$ & $0.130$
& $3.16$ & $98.208$ & $(1.45)\times$ & $0.274$ \\
& {\color{gray!60}\scriptsize$\pm0.53$} & {\color{gray!60}\scriptsize$\pm0.433$} & {\color{gray!60}\scriptsize$\pm0.11$}   & {\color{gray!60}\scriptsize$\pm0.275$}
& {\color{gray!60}\scriptsize$\pm0.10$} & {\color{gray!60}\scriptsize$\pm0.759$} & {\color{gray!60}\scriptsize$\pm0.16$}   & {\color{gray!60}\scriptsize$\pm0.578$} \\
\bottomrule
\end{tabular}%
}
\vspace{0.3em}
\par\noindent{\scriptsize Spd.: LKH speedup vs.\ unpruned base. {\color{gray!60}Standard deviations shown in gray.}}
\end{table}

\vspace{-3mm}
\section{Further Analysis}\label{sec:features_failures}
\vspace{-1mm}

Section~\ref{sec:per_family} decomposed the corridor coverage loss into two layers. The corridor case, though the hardest distribution, reveals the mechanism that makes the two-stage pipeline work. We now zoom in on Layer~2 (the loss from learned pruning), using a single-instance visualisation followed by a family-wide failure-mode classification.

\begin{figure*}[!h]
\centering
\includegraphics[width=0.98\textwidth]{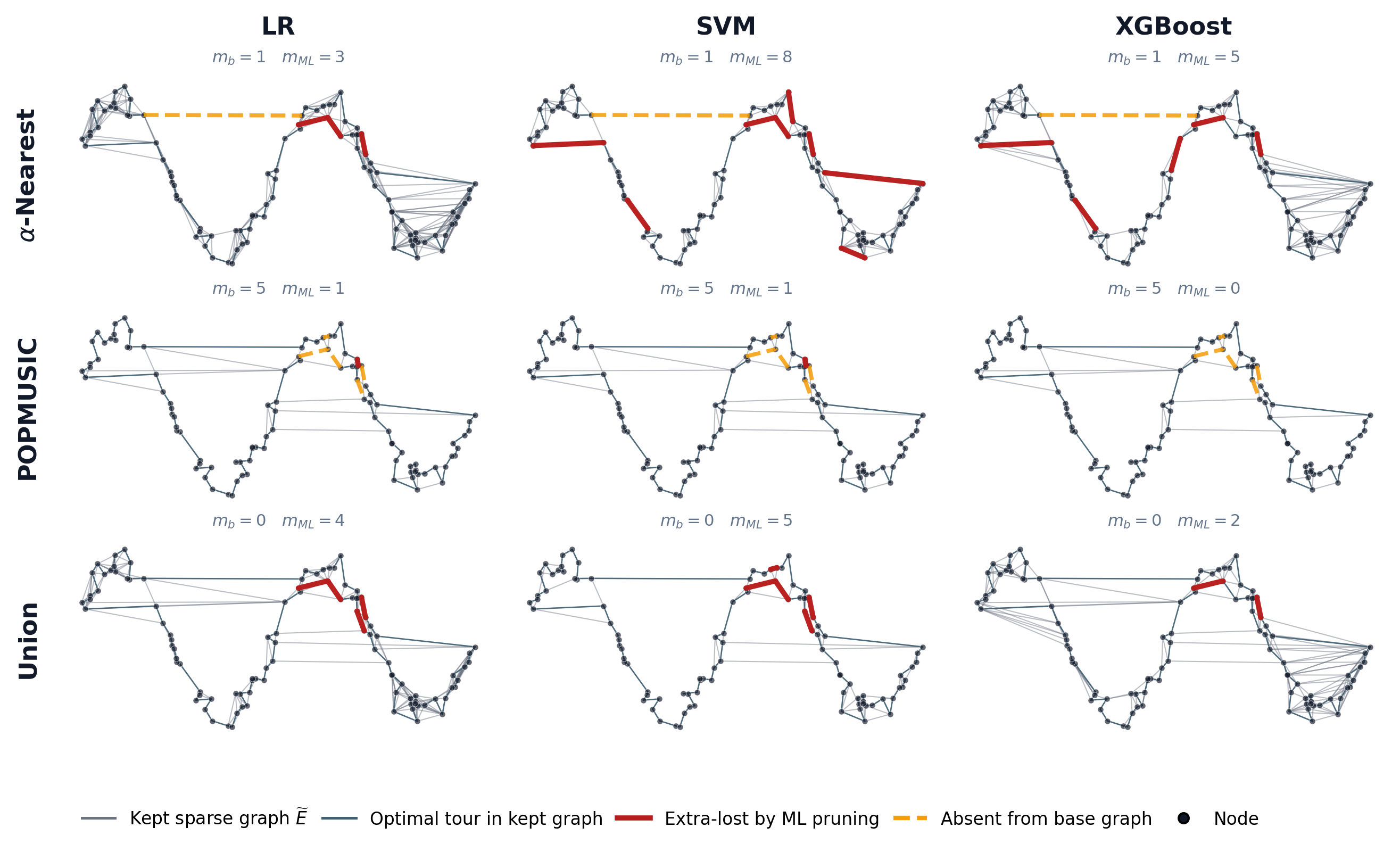}
\vspace{-4mm}
\caption{One MAN\_2D corridor instance, $N{=}100$, pruned by candidate $\times$ model. Solid blue edges form the optimal tour, grey edges the kept sparse graph, red edges are extra-lost by ML pruning, dashed orange edges are absent from the base graph already. $m_b$ counts base misses, $m_{ML}$ counts ML losses.}
\label{fig:lostedge_corridor_case}
\end{figure*}

Figure~\ref{fig:lostedge_corridor_case} shows one MAN\_2D corridor instance. $\alpha$-Nearest misses one optimal edge (a long chord that a five-nearest-neighbour rule cannot reach), and Stage~2 removes 3--8 further edges along the body, with no shared pattern across the three models. POPMUSIC misses five short body edges, and Stage~2 removes 0--1 further edges, consistent with the base graph already having little room for pruning at $1.14$--$2.37N$. The union recovers both kinds of misses, and Stage~2 then removes only 2--5 edges. Crucially, every removed edge is $\alpha$-only (the same moderate-length edges POPMUSIC originally missed); no dual-source or POPMUSIC-only edge is removed by any model. The case-level pattern is that union losses concentrate on the single-source subset where exactly one heuristic endorsed the edge.

\begin{figure*}[!h]
\centering
\includegraphics[width=0.7\textwidth]{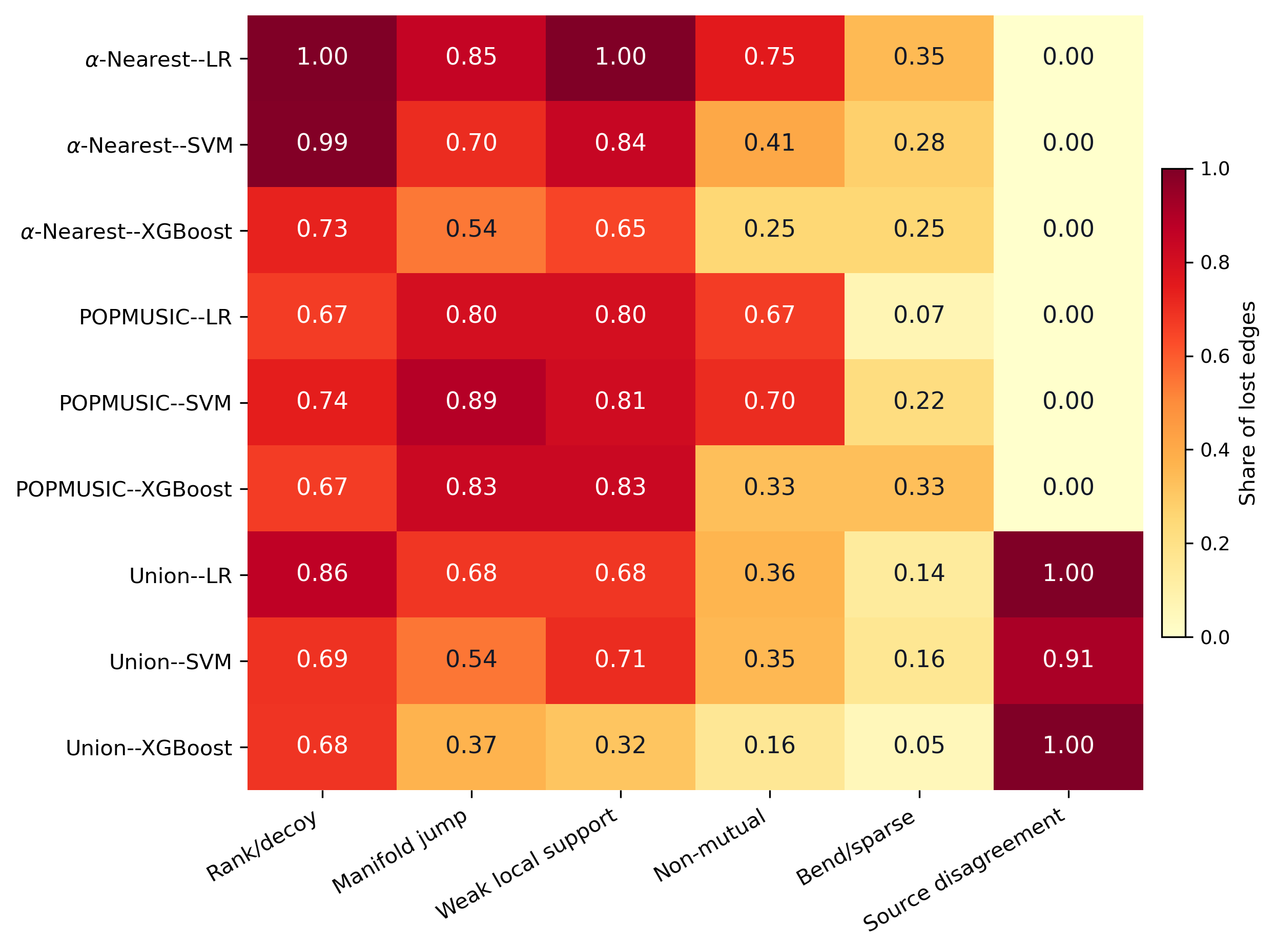}
\vspace{-4mm}
\caption{Share of corridor Stage~2 errors triggering each failure mode, per candidate$\times$model. A mode fires when the relevant feature(s) cross the kept-edge $90\%$ quantile: \emph{Rank/decoy}=high rank, long relative, or many closer decoys; \emph{Manifold jump}=large curve or x-order gap; \emph{Weak local support}=low $k$NN overlap or few common neighbours; \emph{Non-mutual}=non mutual-$k$NN; \emph{Bend/sparse}=near a bend or sparse endpoint; \emph{Source disagreement}=union edge endorsed by only one generator. \emph{Source disagreement} dominates union errors and is absent for single-source bases; \emph{Rank/decoy} and \emph{Weak local support} dominate single-source errors.}
\label{fig:lostedge_corridor_modes}
\end{figure*}

Figure~\ref{fig:lostedge_corridor_modes} extends this to the full corridor test set. On $\alpha$-Nearest and POPMUSIC, the dominant error modes are \emph{Rank/decoy}, \emph{Weak local support}, and \emph{Manifold jump}: the model is left guessing from local geometry alone, and the three classifiers drift visibly. On the union, the story is different: \emph{Source disagreement} is triggered for $100\%$ of the LR and XGBoost mispruned edges and $91\%$ for SVM. In the LR and XGBoost union runs, \emph{not a single dual-source optimal edge is pruned}, while ${\geq}97\%$ of kept edges are dual-source. The union thus reduces the corridor question of which long chord is real to a discrete consensus check, which LR and XGBoost respect perfectly.

\vspace{-3mm}
\section{Conclusions}
\vspace{-2mm}
This paper proposed a two-stage approach to TSP graph sparsification. Stage~1 takes the union of two sparsification heuristics and Stage~2 trains a lightweight classifier to prune the resulting candidate graph. Across four distance types, five spatial distributions, and problem sizes from 50 to 500, the pipeline reduces candidate-graph density by $37$--$47\%$ while retaining ${\geq}99.69\%$ of optimal-tour edges, and yields consistent LKH speedup without increasing the optimality gap.

Three properties explain why this two-stage design works. First, $\alpha$-Nearest and POPMUSIC have complementary failure modes: $\alpha$-Nearest ranks edges by 1-tree structure and misses long chords, while POPMUSIC builds edges from local-tour optima and misses moderate-length edges crossing subproblem boundaries. Their union absorbs both kinds of misses, so Stage~1 alone already achieves ${\geq}99.96\%$ recall. Second, the union annotates every edge with its source provenance. Dual-source edges---those endorsed by both heuristics---are almost always optimal; on the hardest distribution, not a single dual-source edge is mispruned by LR or XGBoost. The learning problem in Stage~2 therefore reduces to identifying which single-source edges to discard, a task simple enough that logistic regression suffices. Third, Stage~1 pre-filters the complete graph from $O(N^2)$ edges to ${\sim}6N$, giving Stage~2 a smaller input with a more balanced class distribution than prior methods that operate on the complete graph.

The method becomes increasingly valuable at larger scales: from $N{=}200$ onward, union + learned pruning achieves higher coverage than POPMUSIC alone. Two natural extensions are theoretical coverage guarantees for learned pruning and replacing the hand-crafted features with a graph neural network, trained on the source-provenance signals that this pipeline exposes.

\bibliographystyle{splncs04}
\bibliography{mlcut_refs}

\end{document}